\documentclass[10pt,twocolumn,letterpaper]{article}

\usepackage{cvpr}
\usepackage{times}
\usepackage{epsfig}
\usepackage{graphicx}
\usepackage{amsmath}
\usepackage{amssymb}
\usepackage[table]{xcolor}
\usepackage{booktabs}

\usepackage[breaklinks=true,bookmarks=false]{hyperref}

\cvprfinalcopy 

\def\cvprPaperID{****} 

\begin{document}

\title{CGNet: A Light-weight Context Guided Network for Semantic Segmentation}

\author{Tianyi Wu$^{1,2}$, Sheng Tang$^{1,}$\thanks{Corresponding author: Sheng Tang (ts@ict.ac.cn)}\, , Rui Zhang$^{1,2}$, Yongdong Zhang$^{1}$\\
$^1$Institute of Computing Technology, Chinese Academy of Sciences, Beijing, China. \\
$^2$ University of Chinese Academy of Sciences, Beijing, China. \\
{\tt\small {\{wutianyi, ts, zhangrui, zhyd\}}@ict.ac.cn}
}

\maketitle


\begin{abstract}
The demand of applying semantic segmentation model on mobile devices has been increasing rapidly. Current state-of-the-art networks have enormous amount of parameters hence unsuitable for mobile devices, while other small memory footprint models follow the spirit of classification network and ignore the inherent characteristic of semantic segmentation. To tackle this problem, we propose a novel Context Guided Network (CGNet), which is a light-weight and efficient network for semantic segmentation. We first propose the Context Guided (CG) block, which learns the joint feature of both local feature and surrounding context, and further improves the joint feature with the global context. Based on the CG block, we develop CGNet which captures contextual information in all stages of the network and is specially tailored for increasing segmentation accuracy. CGNet is also elaborately designed to reduce the number of parameters and save memory footprint. Under an equivalent number of parameters, the proposed CGNet significantly outperforms existing segmentation networks. Extensive experiments on Cityscapes and CamVid datasets verify the effectiveness of the proposed approach. Specifically, without any post-processing and multi-scale testing, the proposed CGNet achieves 64.8\% mean IoU on Cityscapes with less than 0.5 M parameters. The source code for the complete system can be found at \textcolor[rgb]{1,0,0}{https://github.com/wutianyiRosun/CGNet}.
\end{abstract}

\section{Introduction}

\begin{figure}[t]
\centering
\includegraphics[width=\linewidth]{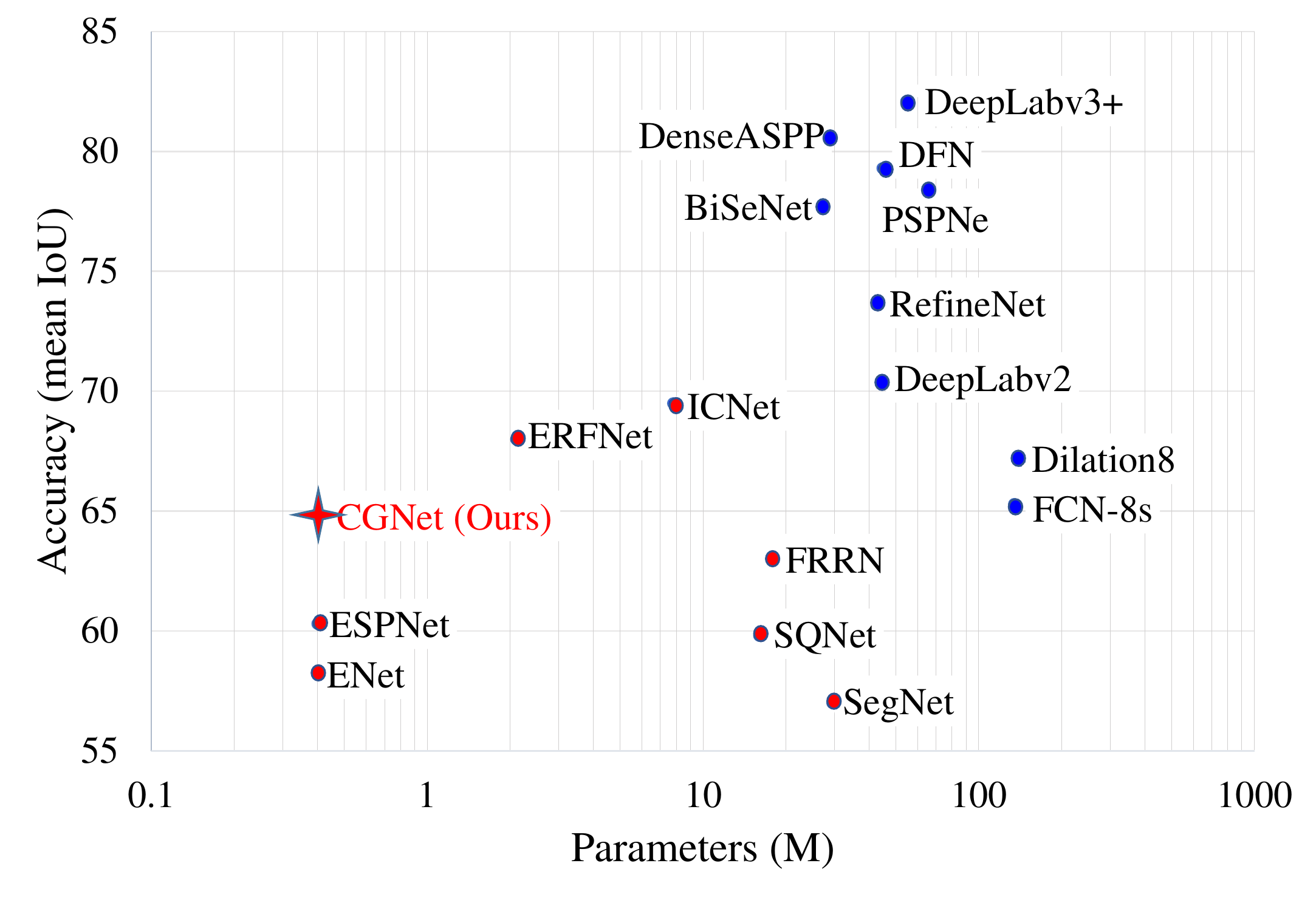}
\caption{Accuracy vs. the number of parameters on Cityscapes \cite{cordts2016cityscapes} \protect\footnotemark[1]. Blue points: high-accuracy methods.  Red points: methods with small memory footprint. Compared with methods of small memory footprint, the proposed CGNet locates in the left-top since it has a lower number of parameters while achieving higher accuracy.}
\label{fig:fig1}
\end{figure}

Recent interest in autonomous driving and robotic systems has a strong demand for deploying semantic segmentation models on mobile devices. It is significant and challenging to design a model with both small memory footprint and high accuracy. Fig.~\ref{fig:fig1} shows the accuracy and the number of parameters of different frameworks on Cityscapes \cite{cordts2016cityscapes} dataset. High-accuracy methods, marked as blue points in Fig.~\ref{fig:fig1}, are transferred from deep image classification networks and have a huge amount of parameters, \eg DFN \cite{Yu_2018_CVPR} of 44.8 M,  DeepLabv3+ \cite{chen2018encoder} of 54.6 M and DenseASPP \cite{yang2018denseaspp} of 28.6 M.
Therefore, most of these high-accuracy methods are unfit for being deployed on mobile devices.
There are some models with small memory footprint, marked as red points in Fig.~\ref{fig:fig1}. Unfortunately, these small footprint methods get low segmentation accuracy, because they only follow the design principle of image classification but ignore the inherent property of semantic segmentation.
To address the above issue, we propose a light-weight network specially tailored for semantic segmentation, named as Context Guided Network (CGNet).


\footnotetext[1]{The methods involved are Dilation8 \cite{yu2015multi}, DeepLabv2 \cite{chen2016deeplab}, SQNet \cite{treml2016speeding}, ENet \cite{paszke2016enet}, PSPNet \cite{Zhao_2017_CVPR}, RefineNet \cite{lin2017refinenet}, FRRN \cite{pohlen2017fullresolution}, FCN-8s \cite{shelhamer2017fully}, SegNet \cite{badrinarayanan2017segnet}, ESPNet \cite{mehta2018espnet}, ERFNet \cite{romera2018erfnet}, ICNet \cite{zhaoicnet}, DenseASPP \cite{yang2018denseaspp}, DeepLabv3+ \cite{chen2018encoder}, DFN \cite{Yu_2018_CVPR}, BiSeNet \cite{yu2018bisenet} and the proposed CGNet.}

\begin{figure}[t]
\centering
\includegraphics[width=\linewidth]{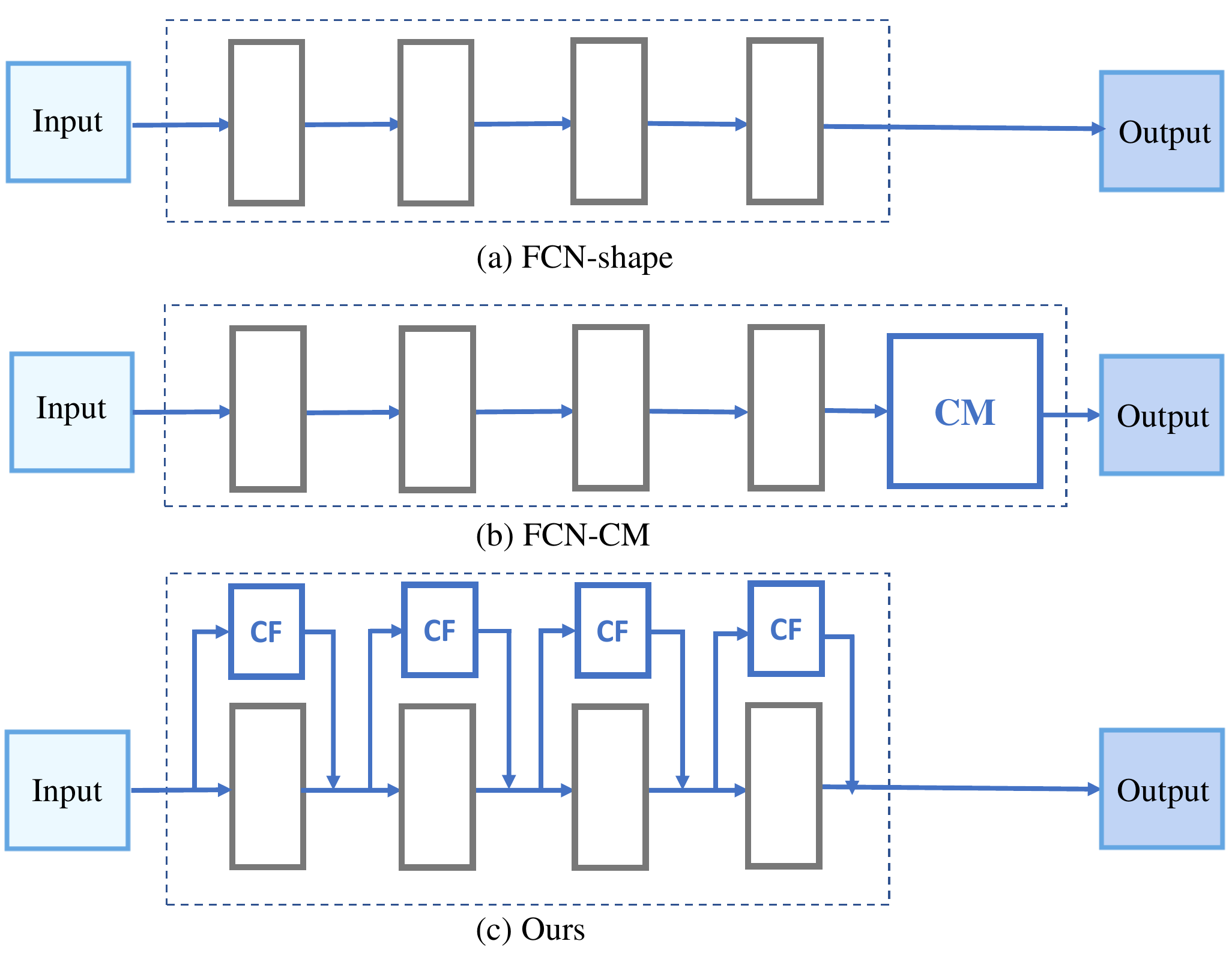}
\caption{Alternative architectures for semantic segmentation. CM: context modules, CF: context features.
(a) FCN-shape models follow the design principle of image classification and ignore contextual information. (b) FCN-CM models only capture contextual information from the semantic level by performing a context module after the encoding stage. (c) The proposed CGNet captures context features in all stages, from both semantic level and spatial level.
}
\label{fig:fig2}
\end{figure}

In order to improve the accuracy, we design a novel CGNet to exploit the inherent property of semantic segmentation.
Spatial dependency and contextual information play an important role to improve accuracy, since semantic segmentation involves both pixel-level categorization and object localization.
Thus, we present Context Guided (CG) block, which is the basic unit of CGNet, to model the spatial dependency and the semantic contextual information effectively and efficiently.
Firstly, CG block learns the joint feature of both local feature and surrounding context. Thus, CG block learns the representation of each object from both itself and its spatially related objects, which contains rich co-occurrence relationship.
Secondly, CG block employs the global context to improve the joint feature. The global context is applied to channel-wisely re-weight the joint feature, so as to emphasize useful components and suppress useless ones.
Thirdly, the CG block is utilized in all stages of CGNet, from bottom to top. Thus, CGNet captures contextual information from both the semantic level (from deep layers) and the spatial level (from shallow layers), which is more fit for semantic segmentation compared with existing methods. Existing segmentation frameworks can be divided into two types: (1) Some methods named FCN-shape models follow the design principle of image classification and ignore the contextual information, \eg ESPNet \cite{mehta2018espnet}, ENet \cite{yu2018bisenet} and FCN \cite{shelhamer2017fully}, as shown in Fig.~\ref{fig:fig2} (a). (2) Other methods named FCN-CM models only capture contextual information from the semantic level by performing context module after the encoding stage, \eg DPC \cite{chen2018searching}, DenseASPP \cite{yang2018denseaspp}, DFN \cite{Yu_2018_CVPR} and PSPNet \cite{Zhao_2017_CVPR}, as shown in Fig.~\ref{fig:fig2} (b). In contrast, the structure of capturing context feature in all stages are more effective and efficient, as shown in Fig.~\ref{fig:fig2} (c). Fourthly, current mainstream segmentation networks have five down-sampling stages which learn too abstract features of objects and missing lots of the discriminative spatial information, causing over-smoothed segmentation boundaries. Differently, CGNet has only three down-sampling stages, which is helpful for preserving spatial information.

Additionally, CGNet is elaborately designed to reduce the number of parameters.
Firstly, it follows the principle of ``deep and thin'' to save memory footprint as much as possible. CGNet only contains 51 layers, and the number of channels in the three stages is 32, 64, 128, respectively. Compared with frameworks \cite{chen2018searching, Yu_2018_CVPR, Zhao_2017_CVPR, yang2018denseaspp} transferred from ResNet \cite{he2016deep} and DenseNet \cite{huang2017densely} which contain hundreds of layers and thousands of channel numbers, CGNet is a light-weighted neural network.
Secondly, to further reduce the number of parameters and save memory footprint, CG block adopts channel-wise convolutions, which removes the computational cost across channels. Finally, experiments on Cityscapes \cite{cordts2016cityscapes} and CamVid \cite{brostow2008segmentation} verify the  effectiveness and efficiency of the proposed CGNet. Without any pre-processing, post-processing, or complex upsampling, our model achieves 64.8\% mean IoU on Cityscapes test set with less than 0.5 M parameters. \emph{We will release the code soon.}

Our main contributions could be concluded as:
\begin{itemize}
\item We analyze the inherent property of semantic segmentation and propose CG block which learns the joint feature of both local feature and surrounding context and further improves the joint feature with global context.
\item We design CGNet, which applies CG block to effectively and efficiently capture contextual information in all stages. The backbone of CGNet is particularly tailored for increasing segmentation accuracy.
\item We elaborately design the architecture of CGNet to reduce the number of parameters and save memory footprint. Under an equivalent number of parameters, the proposed CGNet significantly outperforms existing segmentation networks (e.g., ENet and ESPNet).
\end{itemize}

\begin{figure*}[t]
\centering
\includegraphics[width=\linewidth]{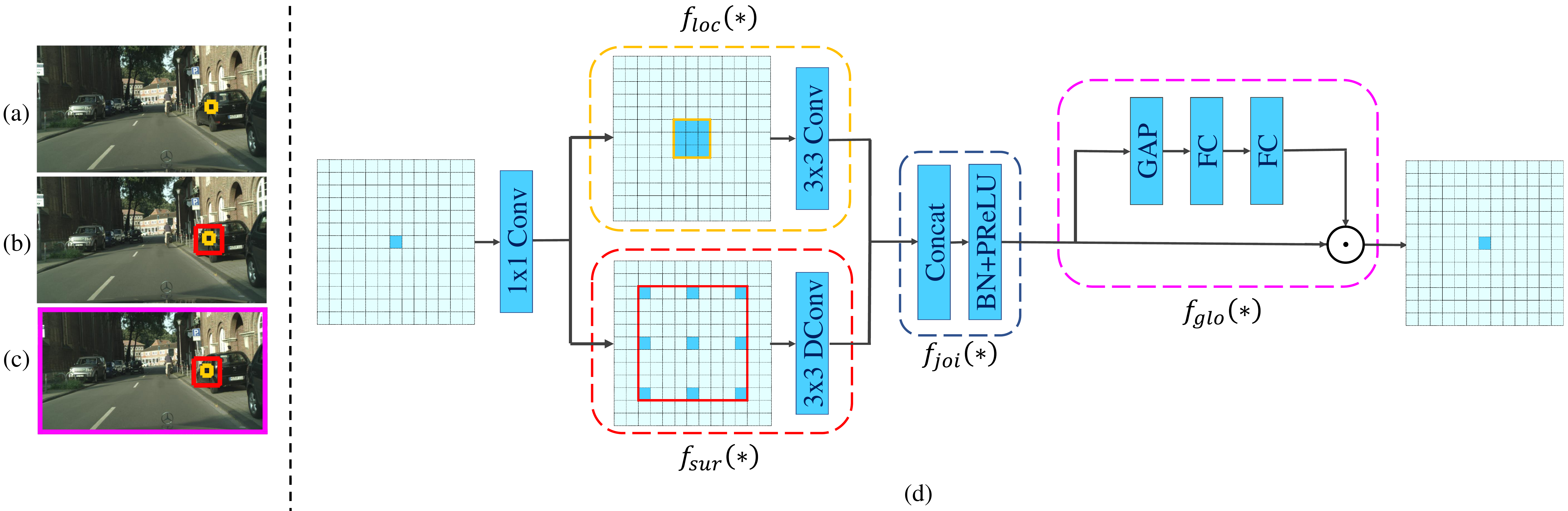}
\caption{An overview of the Context Guided block. (a) It is difficult to categorize the yellow region when we only pay attention to the yellow region itself. (b) It is easier to recognize the yellow region with the help of its surrounding context (red region). (c) Intuitively, we can categorize the yellow region with a higher degree of confidence when we further consider the global contextual information (purple region). (d) The structure of Context Guided block, which consists of local feature extractor $f_{loc}(*)$, surrounding context extractor $f_{sur}(*)$, joint feature extractor $f_{joi}(*)$, and global context extractor $f_{glo}(*)$. ($\cdot$) represents element-wise multiplication.}
\label{fig:fig3}
\end{figure*}

\begin{figure}[t]
\centering
\includegraphics[width=\linewidth]{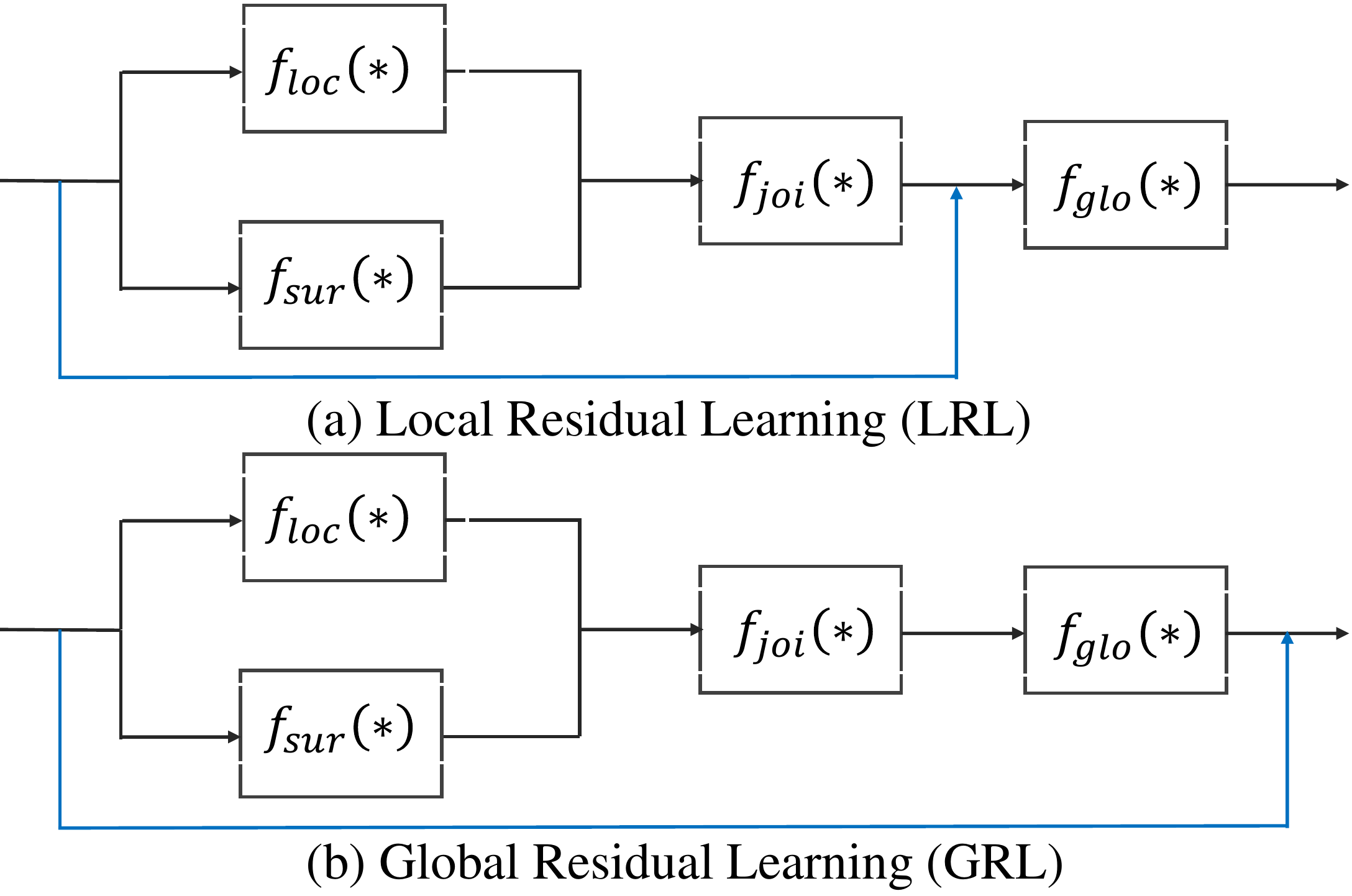}
\caption{Structure of Local Residual Learning (LRL) and Global Residual Learning (GRL).}
\label{fig:fig6}
\vspace{-15pt}
\end{figure}

\section{Related Work}
In this section, we introduce related work on semantic segmentation, including small semantic segmentation models and contextual information models, as well as related work on attention models.

\textbf{Small semantic segmentation models:}\quad
Small semantic segmentation models require making a good trade-off between accuracy and model parameters or memory footprint. ENet \cite{paszke2016enet} proposes to discard the last stage of the model and shows that semantic segmentation is feasible on embedded devices. However,
ICNet \cite{zhaoicnet} proposes a compressed-PSPNet-based image cascade network to speed up the semantic segmentation.
More recent ESPNet \cite{nassar2018deep} introduces a fast and efficient convolutional network for semantic segmentation of high-resolution images under resource constraints.
Most of them follow the design principles of image classification, which makes them have poor segmentation accuracy.

\textbf{Contextual information models:}\quad
Recent works \cite{chen2017rethinking,Ding_2018_CVPR,Yu_2018_CVPR,zhang2018context} have shown that contextual information is helpful for models to predict high-quality segmentation results. One direction is to enlarge the receptive field of filter or construct a specific module to capture contextual information. Dilation8 \cite{yu2015multi} employs multiple dilated convolutional layers after class likelihood maps to exercise multi-scale context aggregation. SAC \cite{zhang2017scale}  proposes a scale-adaptive convolution to acquire flexible-size receptive fields. DeepLab-v3 \cite{chen2017rethinking} employs  Atrous Spatial Pyramid Pooling \cite{chen2016deeplab} to capture useful contextual information with multiple scales. Following this, the work \cite{yang2018denseaspp} introduces DenseASPP to connect a set of atrous convolutional layers for generating multi-scale features. However, the work \cite{zhang2017global} proposes a Global-residual Refinement Network through exploiting global contextual information to predict the parsing residuals. PSPNet \cite{Zhao_2017_CVPR} introduces four pooling branches to exploit global information from different subregions. By contrast, some other approaches directly construct information propagation model. SPN \cite{liu2017learning} constructs a row/column linear propagation model to capture dense and global pairwise relationships in an image, and PSANet \cite{Zhao_2018_ECCV} proposes to learn the adaptively point-wise context by employing bi-directional information propagation. Another direction is to use Conditional Random Field (CRF) to model the long-range dependencies.  CRFasRNN \cite{zheng2015conditional} reformulates DenseCRF with pairwise potential functions and unrolls the mean-field steps as recurrent neural networks, which composes a uniform framework and can be learned end-to-end. Differently, DeepLab frameworks \cite{chen2016deeplab} use DenseCRF \cite{krahenbuhl2011efficient} as post-processing. After that, many approaches combine CRFs and DCNNs in a uniform framework, such as combining Gaussian CRFs \cite{chandra2016fast} and specific pairwise potentials \cite{jampani2016learning}. More recently, CCL \cite{Ding_2018_CVPR} proposes a novel context contrasted local feature that not only leverages the informative context but also spotlights the local information in contrast to the context. DPC \cite{chen2018searching} proposes to search for efficient multi-scale architectures by using architecture search techniques. Most of these works explore context information in the decoder phase and ignore surrounding context, since they take classification network as the backbone of the segmentation model. In contrast, our approach proposes to learn the joint feature of both local feature and surrounding context feature in the encoder phase, which is more representative for semantic segmentation than the feature extracted by the classification network.

\begin{figure*}[t]
\centering
\includegraphics[width=0.95\linewidth]{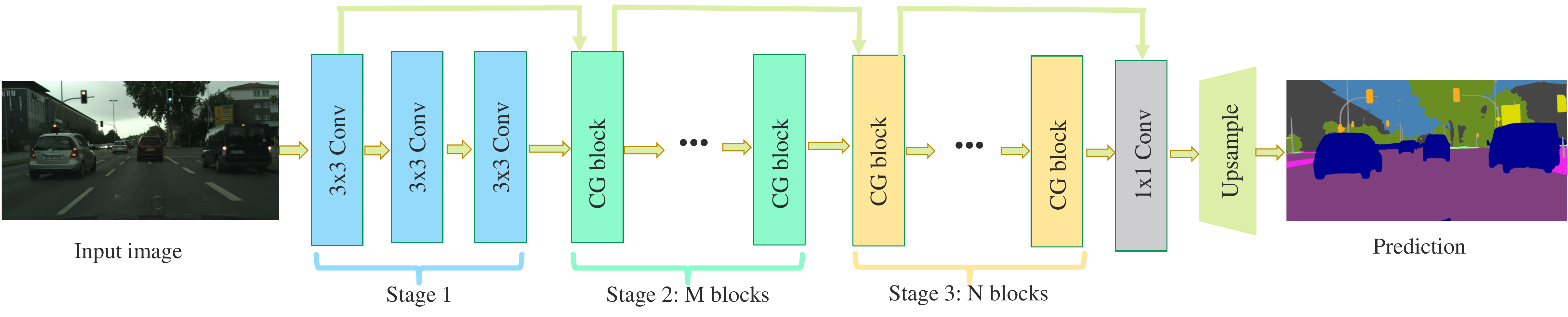}
\caption{The architecture of the proposed Context Guided Network. ``M''and ``N'' are the number of CG blocks in stage 2 and stage 3 respectively.}
\label{fig:fig4}
\vspace{-15pt}
\end{figure*}

\textbf{Attention models:}\quad
Recently, attention mechanism has been widely used for increasing model capability. RNNsearch \cite{bahdanau2014neural} proposes an attention model that softly weighs the importance of input words when predicting a target word for machine translation. Following this, SA \cite{chen2016attention} proposes an attention mechanism that learns to softly weigh the features from different input scales when predicting the semantic label of a pixel. SENet \cite{hu2017squeeze} proposes to recalibrate channel-wise feature responses by explicitly modeling interdependencies between channels for image classification. More recently, NL \cite{wang2018non} proposes to compute the response at a position as a weighted sum of the features at all positions for video classification. In contrast, we introduce the attention mechanism into semantic segmentation. Our proposed CG block uses the global contextual information to compute a weight vector, which is employed to refine the joint feature of both local feature and surrounding context feature.


\section{Proposed Approach}
In this work, we develop CGNet, a light-weight neural network for semantic segmentation on mobile devices. In this section, we first elaborate the important component CG block. Then we present the details of the proposed CGNet. Finally, we compare CG block with similar units.

\subsection{Context Guided Block}

The CG block is inspired by the human visual system, which depends on contextual information to understand the scene.
As shown in Fig.~\ref{fig:fig3} (a), suppose the human visual system tries to recognize the yellow region, which is difficult if we only pay attention to this region itself.
In Fig.~\ref{fig:fig3} (b), we define the red region as the surrounding context of the yellow region. If both the yellow region and its surrounding context are obtained, it is easier to assign the category to the yellow region. Therefore, the surrounding context is helpful for semantic segmentation.
For Fig.~\ref{fig:fig3} (c), if the human visual system further captures the global context of the whole scene (purple region) along with the yellow region and its surrounding context (red region), it has a higher degree of confidence to categorize the yellow region. Therefore, both surrounding context and global context are helpful for improving the segmentation accuracy.

\begin{table}[t]
\begin{center}
\begin{tabular}{|lccc|}
\hline
Name  & {Type} &  Channel & {Output size} \\
\hline\hline
 &  3$\times$3 Conv (stride=2) & 32    & 340 $\times$ 340   \\
stage 1 &  3$\times$3 Conv (stride=1) & 32 & 340 $\times$ 340   \\
&  3$\times$3 Conv (stride=1) & 32  & 340 $\times$340   \\
\hline
stage 2  &  CG block (r=2) $\times$  {M} & 64 &170 $\times$ 170   \\
\hline
stage 3   &  CG block (r=4) $\times$ {N}  & 128  & 85 $\times$ 85   \\
\hline
         & 1$\times$1 Conv(stride=1)  & 19 & 85 $\times$ 85 \\
\hline
\end{tabular}
\end{center}
\caption{The CGNet architecture for Cityscapes. Input size is 3 $\times$ 680 $\times$ 680. ``Conv'' represents the operators of Conv-BN-PReLU. ``r'' is the rate of Atrous/dilated convolution in surrounding context extractor $f_{sur}(*)$. ``M'' and ``N'' are the number of CG blocks in stage 2 and stage 3 respectively.}
\label{table:tab1}
\vspace{-10pt}
\end{table}


Based on the above analysis, we introduce CG block to take full advantage of local feature, surrounding context and global context. CG block consists of  a local feature extractor $f_{loc}(*)$, a surrounding context extractor $f_{sur}(*)$ , a joint feature extractor $f_{joi}(*)$, and a global context extractor $f_{glo}(*)$, as shown in Fig.~\ref{fig:fig3} (d).
CG block contains two main steps.
In the first step,  $f_{loc}(*)$ and $f_{sur}(*)$ is employed to learn local feature and the corresponding surrounding context respectively.
$f_{loc}(*)$ is instantiated as 3 $\times$ 3 standard convolutional layer to learn the local feature from the 8 neighboring feature vectors, corresponding to the yellow region in Fig.~\ref{fig:fig3} (a).
Meanwhile, $f_{sur}(*)$ is instantiated as a 3 $\times$ 3 atrous/dilated convolutional layer since atrous/dilated convolution has a relatively large receptive field to learn the surrounding context efficiently, corresponding to the red region in Fig.~\ref{fig:fig3} (b).
Thus, $f_{joi}(*)$ obtains the joint feature from the output of $f_{loc}(*)$ and $f_{sur}(*)$. We simply design $f_{joi}(*)$ as a concatenation layer followed by the Batch Normalization (BN) and Parametric ReLU (PReLU) operators.
In the second step, $f_{glo}(*)$ extracts global context to improve the joint feature. Inspired by SENet \cite{hu2017squeeze}, the global context is treated as a weighted vector and is applied to channel-wisely refine the joint feature, so as to emphasize useful components and suppress useless one.
In practice, we instantiate $f_{glo}(*)$ as a global average pooling layer to aggregate the global context corresponding to the purple region in Fig.~\ref{fig:fig3} (c), followed by a multilayer perceptron to further extract the global context. Finally, we employ a scale layer to re-weight the joint feature with the extracted global context. Note that the refining operation of $f_{glo}(*)$ is adaptive for the input image since the extracted global context is generated from the input image.


Furthermore, the proposed CG block employs residual learning \cite{he2016deep} which helps to learn highly complex features and to improve gradient back-propagation during training. There are two types of residual connection in the proposed CG block. One is local residual learning (LRL), which connects input and the joint feature extractor $f_{joi}(*)$. The other is global residual learning (GRL),  which bridges input and the global feature extractor $f_{glo}(*)$. Fig.~\ref{fig:fig6} (a) and (b) show these two cases, respectively. Intuitively, GRL has a stronger capability than LRL to promote the flow of information in the network.


\begin{table}[t]
\begin{center}
\begin{tabular}{|lcc|}
\hline
Method  & \bf{$f_{sur}(*)$} & {mIoU (\%)} \\
\hline\hline
CGNet\_M3N15 & No &  54.6 \\
CGNet\_M3N15 & Single & 55.4 \\
CGNet\_M3N15 & Full  &  59.7 \\
\hline
\end{tabular}
\end{center}
\caption{Evaluation results of surrounding context extractor on Cityscapes validation set. Here we set M=3, N=15. }
\label{table:tab2}
\end{table}


\begin{table}[t]
\begin{center}
\begin{tabular}{|lcc|}
\hline
Method  & $f_{glo}(*)$ & {mIoU (\%)} \\
\hline\hline
CGNet\_M3N15 & w/o &58.9   \\
CGNet\_M3N15 & w   & 59.7 \\
\hline
\end{tabular}
\end{center}
\caption{Evaluation results of global context extractor on Cityscapes validation set. Here we set M=3, N=15. }
\label{table:tab3}
\end{table}
\subsection{Context Guided Network}

Based on the proposed CG block, we elaborately design the structure of CGNet to reduce the number of parameters, as shown in Fig.~\ref{fig:fig4}.
CGNet follows the major principle of ``deep and thin'' to save memory footprint as much as possible. Different from frameworks transferred from deep image classification networks which contain hundreds of layers and thousands of channel numbers, CGNet only consists of 51 convolutional layers with small channel numbers.
In order to better preserve the discriminative spatial information, CGNet has only three down-sampling stages and obtains 1/8 feature map resolution, which is much different from mainstream segmentation networks with five down-sampling stages and 1/32 feature map resolution. The detailed architecture of our proposed CGNet is presented in Tab.~\ref{table:tab1}.
In stage 1, we stack only three standard convolutional layers to obtain the feature map of 1/2 resolution, while in stage 2 and 3, we stack $M$ and $N$ CG blocks to downsample the feature map to 1/4 and 1/8 of the input image respectively. For stage 2 and 3, the input of their first layer are gained from combining the first and last blocks of their previous stages, which encourages feature reuse and strengthen feature propagation. In order to improve the flow of information in CGNet, we take the input injection mechanism which additionally feeds 1/4 and 1/8 downsampled input image to stage 2 and stage 3 respectively. Finally, a 1 $\times$ 1 convolutional layer is employed to produce the segmentation prediction.

Note that CG block is employed in all units of stage 2 and 3, which means CG block is utilized almost in all the stages of CGNet. Therefore, CGNet has the capability of aggregating contextual information from bottom to top, in both semantic level from deep layers and spatial level from shallow layers. Compared with existing segmentation frameworks which ignore the contextual information or only capture contextual information from the semantic level by performing context module after the encoding stage, the structure of CGNet is elaborately tailored for semantic segmentation to improve the accuracy.

Furthermore, in order to further reduce the number of parameters, feature extractor $f_{loc}(*)$ and $f_{sur}(*)$ employ channel-wise convolutions, which remove the computational cost across channels and save much memory footprint.
The previous work \cite{howard2017mobilenets} employs 1 $\times$ 1 convolutional layer followed channel-wise convolutions for promoting the flow of information between channels. However, this design is not suitable for the proposed CG block, since the local feature and the surrounding context in CG block need to maintain channel independence. Additional experiments also verify this observation.



\begin{table}[t]
\begin{center}
\begin{tabular}{|lcc|}
\hline
Method  & {Input Injection} & {mIoU (\%)} \\
\hline\hline
CGNet\_M3N15 & w/o &  59.4  \\
CGNet\_M3N15 & w   &  59.7  \\
\hline
\end{tabular}
\end{center}
\caption{The effectiveness of Input Injection mechanism. Here we set M=3, N=15. }
\label{table:tab4}
\end{table}

\begin{table}[t]
\begin{center}

\begin{tabular}{|lcc|}
\hline
Method & Activation & {mIoU (\%)} \\
\hline\hline
CGNet\_M3N15  &  ReLU &  58.1  \\
CGNet\_M3N15  &  PReLU & 59.7   \\
\hline
\end{tabular}
\end{center}
\caption{The effectiveness of ReLU and PReLU. Here we set M=3, N=15. }
\label{table:tab5}
\end{table}

\subsection{Comparision with Similar Works}
ENet unit \cite{paszke2016enet} employs a main convolutional layer to extract single-scale feature, which results in lacking of local features in the deeper layers of the network and lacking surrounding context at the shallow layers of the network. MobileNet unit \cite{howard2017mobilenets} employs a depth-wise separable convolution that factorizes standard convolutions into depth-wise convolutions and point-wise convolutions. Our proposed CG block can be treated as the generalization of MobileNet unit. When $f_{sur}(*)=0$ and $f_{glo}(*)=1$, our proposed CG block will degenerate to MobileNet unit. ESP unit \cite{mehta2018espnet} employs K parallel dilated convolutional kernels with different dilation rates to learn multi-scale features. Inception unit \cite{szegedy2015going} is proposed to approximate a sparse structure and process multi-scale visual information for image classification. CCL unit \cite{Ding_2018_CVPR} leverages the informative context and spotlights the local information in contrast to the context, which is proposed to learn locally discriminative features form block3, block4 and block5 of ResNet-101, and fuse different scale features through gated sum scheme in the decoder phase. In contrast to them, the CG block is proposed to learn the joint feature of both local feature and surrounding context feature in the encoder phase.



\section{Experiments}
In this section, we evaluate the proposed CGNet on Cityscapes \cite{cordts2016cityscapes} and CamVid\cite{brostow2008segmentation}. Firstly, we introduce the datasets and the implementation protocol. Then the contributions of each component are investigated in ablation experiments on Cityscapes validation set. Finally, we perform comprehensive experiments on Cityscapes and CamVid benchmarks and compare with the state-of-the-art works to verify the effectiveness of CGNet.

\begin{table}[t]
\begin{center}
\begin{tabular}{|lccc|}
\hline
 $M$ & $N$  & Parameters (M) &{mIoU (\%)}\\
\hline
3 &  9  &  0.34  & 56.5\\
3 &  12  &  0.38  & 58.1\\
6 &  12  &  0.39 & 57.9\\
3 &  15  & 0.41  & 59.7 \\
6 &  15  & 0.41 & 58.4 \\
3 &  18  & 0.45  & 61.1\\
3 &  21  & 0.49 & 63.5\\
\hline
\end{tabular}
\end{center}
\caption{Evaluation results of CGNet with different M and N on Cityscapes validation set. M: the number of CG blocks in stage 2; N: the number of CG blocks in stage 3. }
\label{table:tab6}
\vspace{-5pt}
\end{table}


\subsection{Experimental Settings}

\paragraph{Cityscapes Dataset}
The Cityscapes dataset contains 5, 000 images collected in street scenes from 50 different cities. The dataset is divided into three subsets, including 2, 975 images in training set, 500 images in validation set and 1, 525 images in testing set. High-quality pixel-level annotations of 19 semantic classes are provided in this dataset. Segmentation performances are reported using the commonly Intersection-over-Union (IoU).

\vspace{-10pt}
\paragraph{CamVid Dataset}
The CamVid is a road scene dataset from the perspective of a driving automobile. The dataset involves 367 training images, 101 validation images and 233 testing images. The images have a resolution of 480 $\times$ 360. The performance is measured by pixel intersection-over-union (IoU) averaged across the 11 classes.
\vspace{-10pt}
\paragraph{Implementation protocol}
 All the experiments are performed on the PyTorch platform with $2\times$ V100 GPU. We employ the ``poly'' learning rate policy, in which we set base learning rate to $0.001$ and power to $0.9$.  For optimization, we use ADAM \cite{kingma2015} with batch size 14, betas=(0.9, 0.999), and weight decay $0.0005$ in training. For data augmentation, we employ random mirror, the mean subtraction and random scale on the input images to augment the dataset during training. The random scale contains $\{0.5, 0.75, 1, 1.5, 1.75, 2.0\}$. The iteration number is set to $60K$ for Cityscapes and CamVid.
For all experiments, we use a single-scale evaluation to compute mean IoU. Our loss function is the sum of cross-entropy terms for each spatial position in the output score map, ignoring the unlabeled pixels.

\subsection{Ablation Studies}

\begin{table}[t]
\begin{center}
\begin{tabular}{|lcc|}
\hline
Method & Residual connections & {mIoU (\%)} \\
\hline\hline
CGNet\_M3N21 & LRL &  57.2  \\
CGNet\_M3N21 & GRL & 63.5   \\
\hline
\end{tabular}
\end{center}
\caption{The effectiveness of local residual learning (LRL) and global residual learning (GRL). Here we set M=3, N=21. }
\label{table:tab7}
\end{table}


\begin{table}[t]
\begin{center}
\begin{tabular}{|lcc|}
\hline
Method &   1x1 Conv & {mIoU (\%)} \\
\hline\hline
CGNet\_M3N21 &   w/  &  53.3 \\
CGNet\_M3N21 & w/o   &  63.5   \\
\hline
\end{tabular}
\end{center}
\caption{The effectiveness of Inter-channel interaction. Here we set M=3, N=21. }
\label{table:tab8}
\vspace{-5pt}
\end{table}

\paragraph{Ablation Study for Surrounding Context Extractor }
We adopt three schemes to evaluate the effectiveness of surrounding context extractor $f_{sur}(*)$. (1) \textbf{No}: CGNet\_M3N15 model does not employ $f_{sur}(*)$, and is configured with the same number of parameters by increasing the number of channels. (2) \textbf{Single}: the surrounding context extractor $f_{sur}(*)$ is employed only in the last block of the framework. (3) \textbf{Full}: the surrounding context extractor $f_{sur}(*)$ is employed in all blocks of the framework. Results are shown in Tab.~\ref{table:tab2}. It is clear that the second and third schemes can improve the accuracy by $ 0.8\%$ and $5.1\%$ respectively, which shows surrounding context is very beneficial for increasing segmentation accuracy and should be employed in all blocks of the framework.
\begin{table*}[t]
\begin{center}
\begin{tabular}{|lcccccc|}
\hline
Method  &  Year & FLOPS (G) $\downarrow $  & Parameters (M) $\downarrow $ &  Memory (M) $\downarrow $  & mIoU (\%) $\uparrow$  &Time (ms) $\downarrow $ \\
\hline\hline
RefineNet \cite{lin2017refinenet}     &'17    & 428.3     & 118.4    &1986.5  &73.6   & $>$1000 \\
PSPNet\_Ms \cite{Zhao_2017_CVPR}          &'17    & 453.6     & 65.6     &2180.6  &78.4   & $>$1000 \\
DenseASP\_Ms \cite{yang2018denseaspp}    &'18    & 214.7     & 28.6     &3997.5  &80.6   & $>$500 \\
\hline
SegNet \cite{badrinarayanan2017segnet} &'15    & 286.0     & 29.5    &  -     & 56.1      & 89.2\\
ENet \cite{paszke2016enet}            &'16    & 3.8       &  0.4    &  -     & 58.3      & 19.3 \\
ERFNet \cite{romera2017efficient}     &'17    & 21.0      & 2.1     &  -     & 68.0      & -  \\
ESPNet \cite{mehta2018espnet}         &'18    & 4.0       &  0.4    &  -     & 60.3      & -   \\
MobileNetV2 \cite{sandler2018mobilenetv2} &'18& 9.1       & 2.1    &  -     & 70.2      & -\\
HRFR \cite{zhang2018high}               &  '18 &-          &-       & -  &74.4      &778.6 \\
\hline
CGNet\_M3N21                          &-      & 6.0       & 0.5    & 334.0    &64.8      & 56.8 \\
\hline
\end{tabular}
\end{center}
\caption{FLOPS, parameter, memory, and accuracy analysis on Cityscapes test set. FLOPS and Memory are estimated for an input of 3$\times$640$\times$360. The running times are computed with input size of 2048$\times$1024. ``-'' indicates the approaches do not report the corresponding results.``Ms'' indicates employing multi-scale inputs with average fusion during testing.
}
\label{table:tab9}
\vspace{-15pt}
\end{table*}

\paragraph{Ablation Study for Global Context Extractor }
We use global context to refine the joint feature learned by $f_{joi}(*)$. As shown in  Tab.~\ref{table:tab3}, global context extractor can improve the accuracy from $58.9\%$ to $59.7\%$, which demonstrates that the global context extractor $f_{glo}(*)$ is desirable for the proposed approach.

\vspace{-10pt}
\paragraph{Ablation Study for the Input Injection Mechanism }
We take input injection mechanism that refers to down-sampling the input image to the resolution of stage 2 and stage 3, and injecting them into the corresponding stage.
As shown in Tab.~\ref{table:tab4}, this mechanism can improve the accuracy from $59.4\% $ to $ 59.7\% $. Intuitively, this performance improvement comes from Input Injection mechanism which increases the flow of information on the network.

\vspace{-10pt}
\paragraph{Ablation Study for Activation Function }
We compare ReLU and PReLU in CGNet\_M3N15, as shown in  Tab.~\ref{table:tab5}. Using PReLU can improve performance from $58.1\% $ to $ 59.7\% $. Therefore, we choose PReLU as the activation function of the proposed model.

\paragraph{Ablation Study for Network Depth}
We train the proposed CGNet with different block nums at each stage and shows the trade-offs between the accuracy and the number of parameters, as shown in Tab.~\ref{table:tab6}. In general, deep networks perform better than shallow ones at the expense of increased computational cost and model size. From Tab.~\ref{table:tab6}, we can find that segmentation accuracy does not increase as M increases when fixing N. For example, we fix $N=12$ and change M from 3 to 6, the mean IoU drops by 0.2 points. So we set $M=3$ (the number of CG blocks in stage 2) for CGNet. Furthermore, we compromise between accuracy and model size by setting different N (the number of CG blocks in stage 3). Our approach achieves the highest mean IoU of 63.5\% on Cityscapes validation set when M=3, N=21.

\vspace{-10pt}
\paragraph{Ablation Study for Residual Learning}
Inspired by \cite{he2016deep}, residual learning is employed in CG block to further improve the information flow. From Tab.~\ref{table:tab7}, compared with LRL, we can find that GRL can improve the accuracy from $57.2\%$ to $63.5\%$. One possible reason is that the GRL has a stronger ability to promote the flow of information in the network, so we choose GRL in the proposed CG block.

\vspace{-10pt}
\paragraph{Ablation Study for Inter-channel Interaction}
Previous work \cite{howard2017mobilenets} employs a 1$\times$1 convolution followed by channel-wise convolution to improve the flow of information between channels and promote inter-channel interaction. Here, We try the 1$\times$1 convolution in CG block but find it damage the segmentation accuracy. As shown in Tab.~\ref{table:tab8}, we can improve the accuracy from $53.3\%$ to $63.5\%$ by removing the 1$\times$1 convolutions. In other words, this interaction mechanism in our CG block hampers the accuracy of our models severely. One possible reason is that the local feature and the surrounding context feature need to maintain channel independence.

\subsection{Comparison with state-of-the-arts}

\paragraph{Efficiency analysis}
Tab.~\ref{table:tab9} reports a comparison of FLOPS (floating point operations), memory footprint, running time and parameters of different models. FLOPS and Memory are estimated with an input size of $3 \times 640 \times 360$, and the running times are computed with an input size of $2048 \times 1024$. According to the $3^{th}$ and $4^{th}$ columns, the FLOPS and Parameters are very close to ENet which is the current smallest semantic segmentation model, yet our method has 6.5\% improvement over ENet. Furthermore, the accuracy of our approach is 4.5\% higher than the very recent model ESPNet \cite{mehta2018espnet} which is based on an efficient spatial pyramid module. With such a few parameters and FLOPS, the proposed CGNet is very suitable to be deployed in mobile devices. Furthermore, compared with deep and state-of-the-art semantic segmentation networks, CGNet\_M3N21 is 131 and 57 times smaller than PSPNet \cite{Zhao_2017_CVPR} and DenseASPP \cite{yang2018denseaspp}, while its category-wise accuracy is only 5.4\% and 5.5\% less respectively. On the other hand, the memory requirement of the proposed model is 334.0 M, which is $10\times$ less than DenseASPP \cite{yang2018denseaspp} (334.0 M vs. 3997.5 M). Finally, we report a comparison of Running times of different model, as shown in the last column in Tab.~\ref{table:tab9}.


\begin{table*}[t]
\begin{center}

\begin{tabular}{|p{4cm}p{2cm}<{\centering}p{2.5cm}<{\centering}p{2.3cm}<{\centering}p{2.5cm}<{\centering}|}
\hline
Method & Pretrain & Parameters (M) $\downarrow $  &mIoU cat (\%) $\uparrow $&mIoU cla (\%) $\uparrow $\\
\hline\hline
SegNet \cite{badrinarayanan2017segnet}       &ImageNet   & 29.5    & 79.1    & 56.1 \\
FCN-8s \cite{shelhamer2017fully}             &ImageNet   & 134.5   & 85.7    & 65.3\\
ICNet \cite{zhaoicnet}                       &ImageNet   & 7.8     &   -  & 69.5 \\

DeepLab-v2+CRF \cite{chen2016deeplab}        &ImageNet  &44.04    & 86.4     & 70.4\\

BiSeNet (Xception) \cite{yu2018bisenet}  &ImageNet  & 145.0   &   -   & 71.4 \\
SAC \cite{zhang2017scale}                  &ImageNet        & -    &90.6          &78.1 \\
BiSeNet (ResNet-18) \cite{yu2018bisenet} &ImageNet  & 27.0     &  -    & 77.7 \\
PSPNet \cite{Zhao_2017_CVPR}            &ImageNet  & 65.7     & 90.6 & 78.4\\
DFN \cite{Yu_2018_CVPR}                  &ImageNet  & 44.8     &  -    &79.3\\
TKCN \cite{wu2018tree}                &ImageNet     &  -        &91.1      &  79.5\\
DenseASPP \cite{yang2018denseaspp}     & ImageNet   &28.6      &  90.7   &80.6\\
OCNet \cite{yuan2018ocnet}              & ImageNet  & 62.5         &-        &81.7 \\
\hline
ENet \cite{paszke2016enet}                 &From scratch &0.4     & 80.4     &58.3 \\
ESPNet \cite{mehta2018espnet}              &From scratch &0.4     & 82.2     &60.3 \\
FRRN \cite{pohlen2017fullresolution}      & From scratch & 17.7   &   -   & 63 \\

\hline
CGNet\_M3N21                         & From scratch & 0.5  & 85.7     &64.8\\
\hline
\end{tabular}
\end{center}
\caption{Accuracy comparison of our method against other small or high-accuracy semantic segmentation methods on Cityscapes test set, only training with the fine set. `Pretrain'' refers to the models that have been pretrained using external data like ImageNet, and ``-'' indicates that the approaches do not report the corresponding results.}
\label{table:tab10}
\vspace{-10pt}
\end{table*}


\begin{table}[t]
\begin{center}

\begin{tabular}{|lccc|}
\hline
Method           &Year  & Parameters (M) $\downarrow $  & mIoU (\%) $\uparrow $ \\
\hline\hline
SegNet \cite{badrinarayanan2017segnet}  &'15     & 29.5             & 55.6 \\
ENet \cite{paszke2016enet}              &'16     & 0.4              & 51.3 \\
G-FRNet \cite{amirul2017gated}          &'17     &-                 & 68.0\\
BiSeNet \cite{yu2018bisenet}            &'18     & 27.0             & 68.7 \\
\hline
FCN (Res101)                           & -       & 56.8             &  67.5   \\
CGNet\_M3N21                           & -       & 0.5              & 65.6\\
\hline
\end{tabular}
\end{center}
\caption{Accuracy comparison of our method against other semantic segmentation methods on Camvid test set.  ``-'' indicates that the approaches do not report the corresponding results.}
\label{table:tab11}
\vspace{-15pt}
\end{table}

\begin{figure}[t]
\centering
\includegraphics[width=\linewidth]{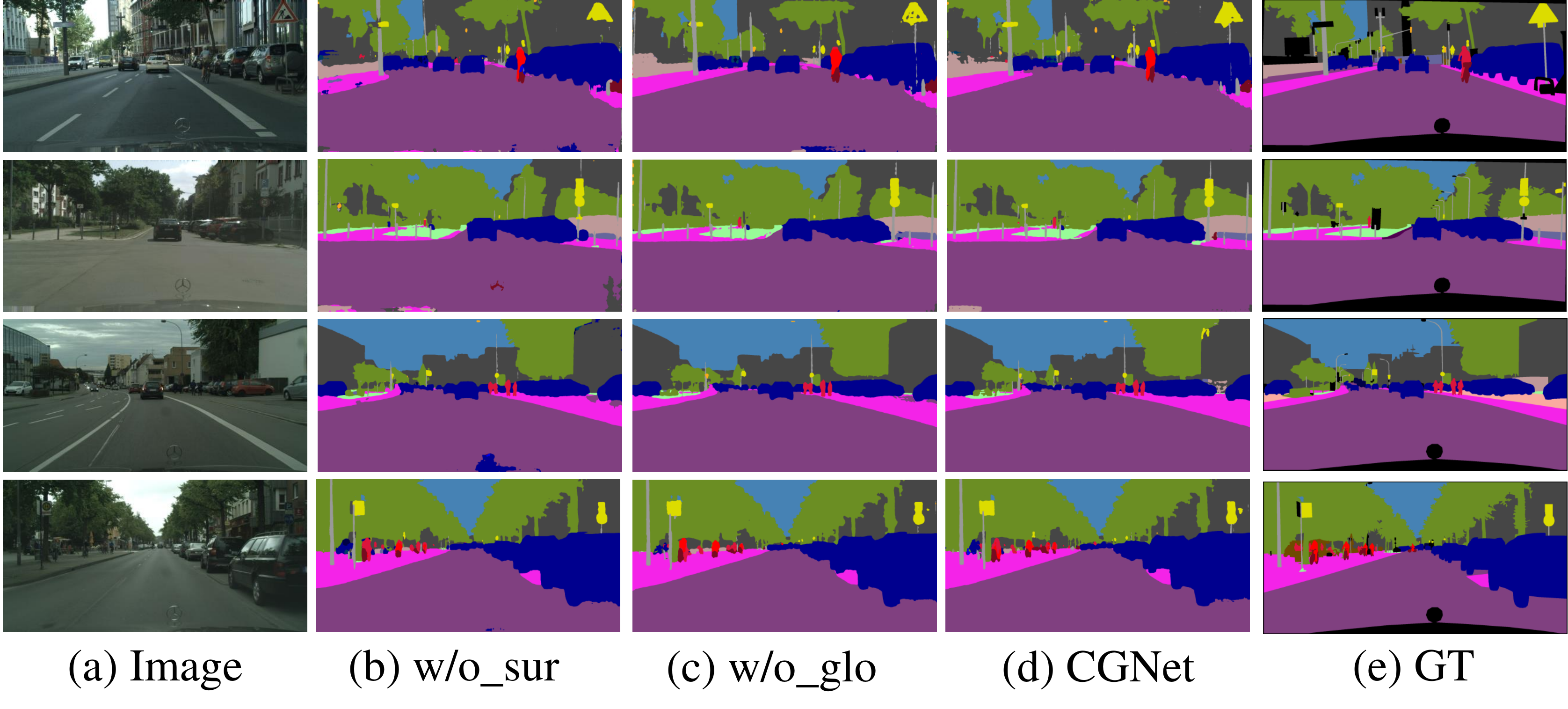}
\caption{Result illustration of CGNet on Cityscapes validation set. From left to right: Input image, prediction of CGNet\_M3N21 without surrounding context feature extractor $f_{sur}(*)$, prediction of CGNet\_M3N21 without global context feature extractor $f_{glo}(*)$, prediction of CGNet\_M3N21 and ground-truth.}
\label{fig:fig5}
\vspace{-15pt}
\end{figure}

\vspace{-10pt}
\paragraph{Accuracy analysis}
We report the evaluation results of the proposed CGNet\_M3N21 on Cityscapes test set and compare to other state-of-the-art methods in Tab.~\ref{table:tab10}. Without any pre-processing, post-processing, or any decoder modules (such as ASPP \cite{chen2016deeplab}, PPM \cite{Zhao_2017_CVPR}), our CGNet\_M3N21 achieves $64.8\%$ 
 in terms of mean IoU (only training on fine annotated images). Note that we do not employ any testing tricks, like multi-scale or complex upsampling. We list the number of model parameters and the segmentation accuracy in Tab.~\ref{table:tab10}. Compared with the methods that do not require pretraining on ImageNet, our CGNet\_M3N21 achieves a relatively large accuracy gain. For example, the mean IoU of proposed CGNet\_M3N21 is about 6.5\%  and 4.5\% higher than ENet \cite{paszke2016enet} and ESPNet \cite{mehta2018espnet} with almost no increase of the model parameters. Besides, it is even quantitatively better than the methods that are pretrained on ImageNet without consideration of memory footprint and speed, such as SegNet \cite{badrinarayanan2017segnet}, and the model parameters of CGNet\_M3N21 is about 60 times smaller than it. We visualize some segmentation results on the validation set of Cityscapes in Fig.~\ref{fig:fig5}.
Tab.~\ref{table:tab11} shows the accuracy result of the proposed CGNet\_M3N21 on CamVid dataset. We use the training set and validation set to train our model. Here, we use 480$\times$360 resolution for training and evaluation. The number of parameters of CGNet\_M3N21 is close to the current smallest semantic segmentation model ENet \cite{paszke2016enet}, and the accuracy of our method is 14.3\% higher than it. The proposed CGNet is just lower 1.9\% than FCN (Res101) with output stride of 8, and the parameters of CGNet is $110 \times$ less than it (0.5 M vs. 56.8 M), which further verifies the effectiveness and efficiency of our method.

\section{Conclusions}
In this paper, we rethink semantic segmentation from its characteristic which involves image recognition and object localization. Furthermore, we propose a novel Context Guided block for learning the joint feature of both local feature and the surrounding context. Based on Context Guided block, we develop a light-weight Context Guided Network for semantic segmentation, and our model allows very memory-efficient inference, which significantly enhances the practicality of semantic segmentation in real-world scenarios. Experiments on the Cityscapes and CamVid show that the proposed CGNet provides a general and effective solution for achieving high-quality segmentation results under the case of resource limited.

{\small
\bibliographystyle{ieee}
\bibliography{egbib}
}

\end{document}